\documentclass[conference]{IEEEtran}
\IEEEoverridecommandlockouts

\usepackage{booktabs}
\usepackage{multirow}
\usepackage[table,xcdraw]{xcolor}
\usepackage{soul}
\usepackage[normalem]{ulem}
\useunder{\uline}{\ul}{}
\usepackage{stackengine}
\usepackage{amsmath}
\usepackage{amsfonts}
\usepackage{listings}
\usepackage{booktabs}
\usepackage{lineno,hyperref}
\usepackage{orcidlink}

\usepackage{cite}
\usepackage{amsmath,amssymb,amsfonts}
\usepackage{algorithmic}
\usepackage{graphicx}
\usepackage{textcomp}
\usepackage{xcolor}
\def\BibTeX{{\rm B\kern-.05em{\sc i\kern-.025em b}\kern-.08em
    T\kern-.1667em\lower.7ex\hbox{E}\kern-.125emX}}
\begin{document}

\title{CogniDual Framework: Self-Training Large Language Models within a Dual-System Theoretical Framework for Improving Cognitive Tasks}

\author{
    \IEEEauthorblockN{
        \textbf{Yongxin Deng}\textsuperscript{1}$^,$\IEEEauthorrefmark{1},
       \textbf{Xihe Qiu}\textsuperscript{1}$^,$\IEEEauthorrefmark{1}$^,$\IEEEauthorrefmark{2},
        \textbf{Xiaoyu Tan}\textsuperscript{2}$^,$\IEEEauthorrefmark{1},
        \textbf{Chao Qu}\textsuperscript{2},
        \textbf{Jing Pan}\textsuperscript{3},\\
        \textbf{Yuan Cheng}\textsuperscript{2},
        \textbf{Yinghui Xu}\textsuperscript{4},
        and \textbf{Wei Chu}\textsuperscript{2}}
    \IEEEauthorblockA{\textsuperscript{1}School of Electronic and Electrical Engineering, Shanghai University of Engineering Science, Shanghai, China\\
                      \textsuperscript{2}INF Technology (shanghai) Co., Ltd., Shanghai, China\\
                      \textsuperscript{3}School of Art, Design and Architecture, Monash University, Melbourne, Australia\\
                      \textsuperscript{4}Artificial Intelligence Innovation and Incubation Institute, Fudan University, Shanghai, China}
     
    \IEEEauthorblockA{\IEEEauthorrefmark{1}This is to indicate the equal contribution.\\
        \IEEEauthorrefmark{2}This is to indicate the corresponding author. Email: qiuxihe@sues.edu.cn}
}

\maketitle

\begin{abstract}
    Cognitive psychology investigates perception, attention, memory, language, problem-solving, decision-making, and reasoning. Kahneman's dual-system theory elucidates the human decision-making process, distinguishing between the rapid, intuitive System 1 and the deliberative, rational System 2. Recent advancements have positioned large language Models (LLMs) as formidable tools nearing human-level proficiency in various cognitive tasks. Nonetheless, the presence of a dual-system framework analogous to human cognition in LLMs remains unexplored. This study introduces the \textbf{CogniDual Framework for LLMs} (CFLLMs), designed to assess whether LLMs can, through self-training, evolve from deliberate deduction to intuitive responses, thereby emulating the human process of acquiring and mastering new information. Our findings reveal the cognitive mechanisms behind LLMs' response generation, enhancing our understanding of their capabilities in cognitive psychology. Practically, self-trained models can provide faster responses to certain queries, reducing computational demands during inference.
\end{abstract}


\section{Introduction}

Cognitive psychology seeks to elucidate the processes by which humans acquire, retain, and retrieve knowledge \cite{evans2013dual,stanovich2008relative}. Kahneman's dual-system theory \cite{kahneman2011thinking, couldry2020costs,guest2021computational, pertwee2022epidemic,balland2022reprint,post2020advancing} emerges as a seminal framework within this realm, offering a nuanced understanding of cognitive operations. This theory outlines two distinct cognitive systems: System 1, which is instinctual and facilitates rapid decision-making with minimal cognitive effort, and System 2, which is methodical and requires deliberate focus for complex reasoning tasks.

In the realm of artificial intelligence, the advent of deep learning and the influx of extensive datasets have precipitated the swift advancement of language models (LMs). These models, particularly those utilizing the Transformer architecture \cite{vaswani2017attention} such as GPT-4 \cite{achiam2023gpt}, have garnered significant attention for their advanced language processing abilities \cite{ouyang2022training, wang2022self, taori2023alpaca, deng2024promoting}, achieving near-human proficiency across numerous linguistic tasks. Trained on expansive natural language corpora, these models demonstrate an ability to comprehend and produce symbol sequences with an intuition akin to human System 1's pattern processing. Furthermore, when prompted to employ CoT problem-solving, LLMs exhibit deep reasoning capabilities paralleling human System 2 \cite{wei2022chain,wang2022self,kojima2022large}. Nevertheless, the persistence of such efficient and accurate outputs in the absence of CoT remains uncertain. Should LLMs achieve this, it would suggest the integration of an intuitive operational process comparable to human System 1.

\textit{Can LLMs internalize System 2's complex reasoning into System 1's intuitive responses through iterative training?} We hypothesize that LLMs, by mimicking human rapid skill acquisition, can generate fast, intuitive answers without additional training data, thus enhancing resource efficiency and reducing dependence on chains of thought (CoT). Our methodology was straightforward yet robust: we commenced by prompting the model with specific reasoning questions, both with and without CoT cues, and subsequently assessed the accuracy of the responses generated under each condition. Following this, the model employed CoT as a scaffold to reengineer non-CoT responses. Theoretically, this self-editing could facilitate the internalization of CoT reasoning steps, potentially enhancing the model's future problem-solving precision without explicit CoT prompts. The final phase involved reevaluating the model's performance post self-improvement to determine if there was an enhancement in CoT-independent operations. Our experiments are designed to uncover whether LLMs can emulate the human cognitive system by internalizing complex reasoning processes, particularly when functioning without direct reasoning instructions.

We applied our methodology to the Vicuna and Llama2 models of varying sizes and evaluated their performance enhancements on reasoning datasets such as GSM8K, ReClor, and LogiQA 2.0. Our findings indicate that LLMs display marked discrepancies in response accuracy when utilizing CoT compared to when it is absent. Following a period of self-training, LLMs exhibited a substantial increase in response precision in scenarios devoid of CoT. This suggests that LLMs are capable of developing intuitive response mechanisms akin to the human cognitive System 1, as well as the deliberate, sequential reasoning characteristic of System 2. Our research demonstrates the potential to cultivate LLMs' System 2 skills into System 1 proficiencies, enabling rapid application.

This paper presents three principal contributions:

\begin{itemize}

    \item We propose a self-iterative framework for large models, which is used to explore whether large models themselves possess characteristics similar to human cognitive structures. 
    \item We demonstrate that LLMs can simulate the dual-system characteristics of human cognition. Through experimentation, we have verified that LLMs can not only perform complex reasoning tasks under the guidance of CoT (similar to human System 2), but also respond relying on pattern recognition intuition without CoT (similar to human System 1). 
    \item We propose and validate a new method that may allow LLMs to maintain efficient and accurate outputs without relying on CoT. This suggests that LLMs can handle tasks in a manner closer to human System 1, and this method is more efficient in terms of computational resources and time because it avoids additional training data or steps, thus it is expected to play a significant role in resource-limited application scenarios.

\end{itemize}

\section{CogniDual Framework}

\subsection{Model Self-Iteration}
\label{model}

\begin{figure}[htbp]

\centerline{\includegraphics[width=0.48\textwidth]{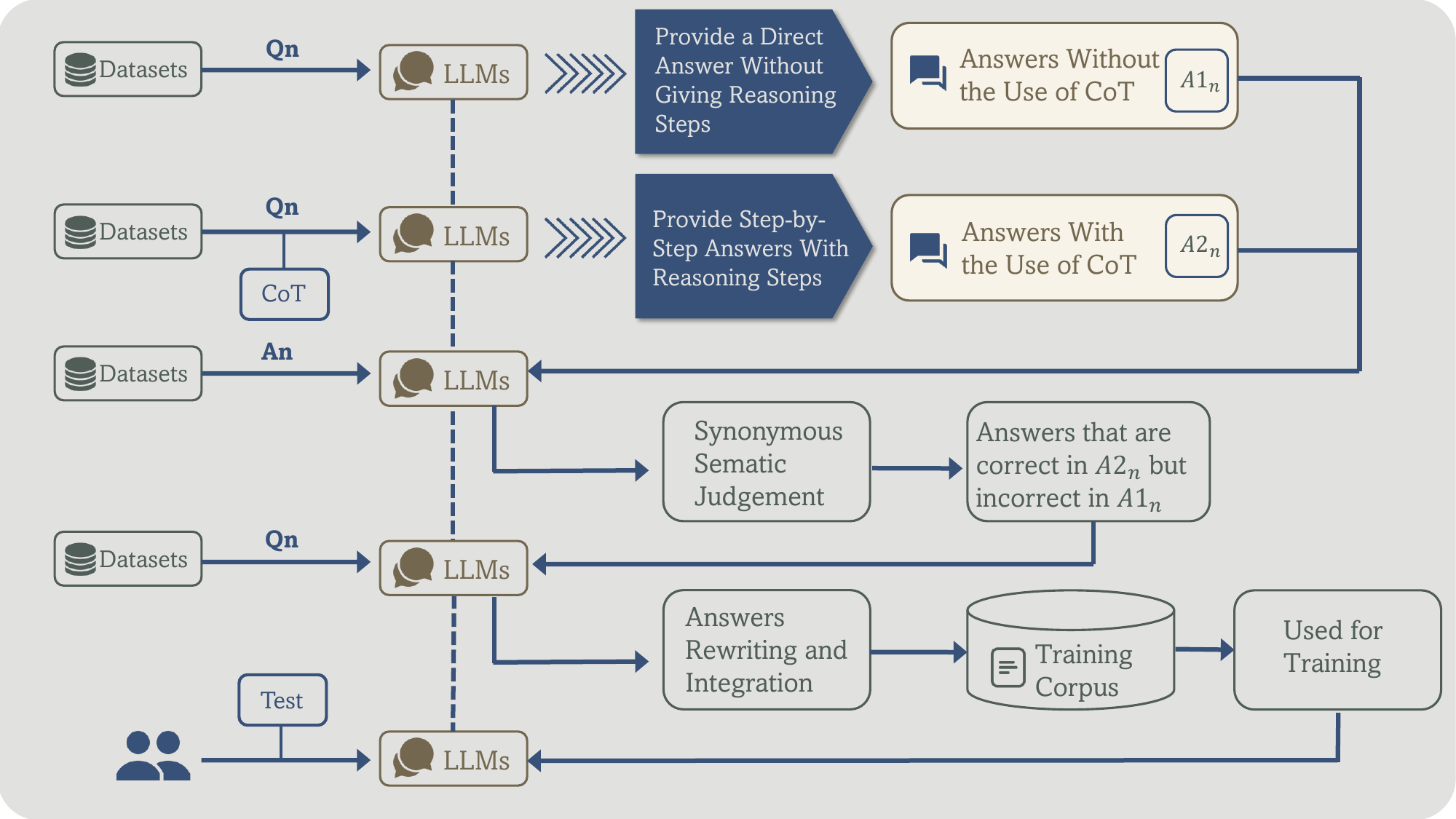}}

\caption{Structural illustration of the CogniDual Framework.}

  \label{fig:main}
\end{figure}

Our CogniDual framework replicates the human learning curve, as depicted in Figure \ref{fig:main}. Initially, we prompt untrained LLMs to answer questions from reasoning datasets without CoT instructions, even compelling the LLMs to provide immediate answers without rationale. We designate this question set as $Q_{n}=\{q_{i},\;\mathrm{for}\; i=1,2,\dots,n \}$, with $n$ denoting the total number of questions. The corresponding answer set is labeled $A1_{n} = \{a1_{i},\;\mathrm{for}\; i=1,2,\dots,n \}$, symbolizing the LLMs’ initial responses, akin to human cognitive System $1$. These question-answer pairs are preserved. 

Subsequently, we introduce CoT directives, guiding LLMs to derive correct answers sequentially. The resulting answers are categorized as $A2_{n}=\{a2_{i},\;\mathrm{for}\; i=1,2,\dots,n \}$. We also maintain these pairs. In the third phase, we furnish the LLMs with standard answers from the dataset, denoted as $A_{n}=\{a_{i},\;\mathrm{for}\; i=1,2,\dots,n\}$.

To quantify the proficiency of our LLMs in responding to $Q_{n}$, we define the accuracy metric as follows:
\begin{equation}
\label{eq:accuracy1}
\text{Acc}(A1_{n}, A_{n}) = \frac{1}{n} \sum_{i=1}^{n} \text{SemanticMatch}(a1_{i}, a_{i}),
\end{equation}
\begin{equation}
\label{eq:accuracy2}
\text{Acc}(A2_{n}, A_{n}) = \frac{1}{n} \sum_{i=1}^{n} \text{SemanticMatch}(a2_{i}, a_{i}),
\end{equation} where $\text{SemanticMatch}(\cdot, \cdot)$ assesses the semantic similarity between the LLM's initial response and the standard answer.
Given that standard answers usually include comprehensive reasoning and do not conform to a 'yes or no' format, we cannot rely on character matching scripts to evaluate the LLMs’ responses. Instead, we engage the LLMs in semantic synonymy judgments to assess the accuracy of $A1_{n}$ and $A2_{n}$ against $A_{n}$, specifically identifying instances where $A2_{n}$ is accurate, and $A1_{n}$ is not. 

The fourth stage involves the LLMs consolidating the correct answers from $A2_{n}$ and the incorrect ones from $A1_{n}$ into new question-answer pairs. Given that $A2_{n}$ responses encompass extensive reasoning, we require LLMs to distill these answers, converting them from elaborate, reasoned responses to concise answers. The specifics of this process will be explored in Section \ref{distillation}. In the final step, we employ these restructured question-answer pairs as training material for LLMs and subsequently assess the LLMs' reasoning capabilities on different questions within the same dataset without CoT. Our experimental approach, chosen for its minimal computational resource demands and deployability, utilizes the LoRA training method \cite{hu2021lora} for LLMs.

\subsection{Pre-training Model Distillation}
\label{distillation}
The framework outlined in Section \ref{model} enables LMs to self-train independently of external interaction. Nevertheless, our framework necessitates that models complete two supplementary tasks when addressing dataset questions: \textbf{1. Synonymy Semantic Judgment:} LMs must determine if $A1_{n}$, $A2_{n}$, and $A_{n}$ are semantically equivalent to assess reasoning accuracy both with and without the CoT; \textbf{2. Answer Rewriting:} Recognizing the impracticality of manually rewriting numerous answers containing reasoning processes into concise responses, we expect LMs to autonomously perform answer rewriting. Suppose we have an open-source LLM denoted by $p_\theta$, which is parameterized by $\theta$. To perform synonymy semantic judgment and achieve $\text{SemanticMatch}$, we can design a specific prompt, $\text{Prompt}_{Semantic}$. This prompt will evaluate whether the two responses $a_i$ and $a_j$ have identical semantic meanings:
\begin{equation}
    \text{SemanticMatch}(a_i,a_j) = p_\theta(a_j,a_i|\text{Prompt}_{Semantic}).
\end{equation} 
For answer rewriting, which is also a common task for chat base LLM, we can design a $\text{Prompt}_{Rewrite}$ to align the generated answer $a_j$ to the target answer $a_i$ with the identical format, and acquire the updated answer $a^\prime_i$:
\begin{equation}
    a^\prime_i = p_\theta(a_j,a_i|\text{Prompt}_{Rewrite}).
\end{equation}

While large-scale models readily accomplish these tasks, smaller models, such as the Llama2-7B, may find them challenging \cite{clavie2023large}. A model that struggles to understand standard answers is unlikely to enhance its capabilities from System 2 to System 1 through self-training alone. To improve training outcomes, we advocate for the pre-training of smaller models using knowledge distillation, equipping them with essential skills for synonymy semantic judgment and answer rewriting. 

Knowledge distillation \cite{huang2022context,agarwal2023gkd,huang2022context,li2022explanations,ho2022large,fu2023specializing,shridhar2023distilling} is a technique for transferring knowledge from a large, complex `teacher' model to a smaller, simpler `student' model, facilitating deployment in resource-limited settings without greatly impacting performance. We employ a simplified approach akin to \textbf{Distilling step-by-step \cite{hsieh2023distilling}}. For synonymy semantic judgment, smaller models generate sample $A1_{n}$, $A2_{n}$, and $A_{n}$, after which GPT-3.5's advanced generative power yields precise judgments and comprehensive explanations. By creating multiple explanations per question, we ensure a clear delineation of the reasoning pathway. GPT-3.5 also supplies sample rewrites and their justifications for the answer rewriting task. Subsequently, smaller model $\theta$ can be trained through supervised fine-tuning using the larger models' outputs $A^\prime$, preparing them for self-improvement and independent practice:
\begin{equation}
    \min_\theta -\mathbb{E}_{(q_i, a^\prime_i) \sim A^\prime}\left[\log p_\theta(a^\prime_i|q_i)\right].
\end{equation}

\section{Experiment}

\begin{table*}[htbp]

\caption{Performance evaluation of diverse LLMs by type and size across various datasets: a detailed assessment with methodological comparisons.}
\begin{center}
    
\resizebox{0.9\textwidth}{!}{%
\begin{tabular}{ccccccc}
\toprule[1.5pt]
                                  &                                     & \multicolumn{5}{c}{\textbf{Models}}                                                         \\ 
\multirow{-2}{*}{\textbf{Method}} & \multirow{-2}{*}{\textbf{Datasets}} & LLAMA2-7B     & LLAMA2-13B    & Vicuna-7B                   & Vicuna-13B    & Vicuna-30B    \\ \midrule[0.8pt]
                                  & GSM8K(Acc)                          & 14.6          & 27.9          & 12.1                        & 15.7          & 35.6          \\
                                  & Reclor(Acc)                         & 34.1          & 59.9          & 41.3                        & 62.5          & 75.2          \\
\multirow{-3}{*}{CoT Not Applied} & LogiQA2.0(Acc)                      & 0.5           & 5.3           & 0.1                         & 5.1           & 27.6          \\ \midrule[0.8pt]
                                  & GSM8K(Acc)                          & {\ul 16.8}    & {\ul 29.1}    & {\ul 14.5}                  & {\ul 16.1}    & {\ul 36.7}    \\
                                  & Reclor(Acc)                         & {\ul 51.5}    & {\ul 89.4}    & {\ul 55.1}                  & {\ul 90.8}    & {\ul 99.4}    \\
\multirow{-3}{*}{CoT Applied}     & LogiQA2.0(Acc)                      & {\ul 77.3}    & {\ul 97.9}    & {\ul 79.4}                  & {\ul 98.4}    & {\ul 99.3}    \\ \midrule[0.8pt]
                                  & GSM8K(Acc)                          & 14.9          & 28.4          & 12.2                        & 16.1          & 35.7          \\
                                  & Reclor(Acc)                         & 35            & 65            & 41.8                        & 65.6          & 80.3          \\
\multirow{-3}{*}{CFLLMs(10)}      & LogiQA2.0(Acc)                      & 2.8           & 12.2          & 1.2                         & 9.2           & 52.7          \\ \midrule[0.8pt]
                                  & GSM8K(Acc)                          & 14.9          & 28.6          & {\color[HTML]{FE0000} 11.9} & 15.8          & 35.7          \\
                                  & Reclor(Acc)                         & 37.2          & 72.3          & 43.8                        & 72            & 98.6          \\
\multirow{-3}{*}{CFLLMs(100)}     & LogiQA2.0(Acc)                      & 9.4           & 33.1          & 7.3                         & 29.1          & 92.1          \\ \midrule[0.8pt]
                                  & GSM8K(Acc)                          & 15.1          & 28.7          & 12.2                        & 16            & 35.9          \\
                                  & Reclor(Acc)                         & 39.8          & 76            & 45.5                        & 78.3          & 98.9          \\
\multirow{-3}{*}{CFLLMs(500)}     & LogiQA2.0(Acc)                      & 9.7           & 72.7          & 8.7                         & 64.4          & 96.3          \\ \midrule[0.8pt]
                                  & GSM8K(Acc)                          & \textbf{15.6} & \textbf{29}   & 12.6                        & 15.9          & \textbf{36.2} \\
                                  & Reclor(Acc)                         & 49            & 77.7          & \textbf{53.4}               & \textbf{78.7} & \textbf{99.2} \\
\multirow{-3}{*}{CFLLMs(1000)}    & LogiQA2.0(Acc)                      & \textbf{18.1} & \textbf{76.2} & 16.7                        & 66.3          & \textbf{96.9} \\ \bottomrule[1.5pt]
\end{tabular}

}
\end{center}

\label{table1}
\end{table*}

\subsection{Experimental Objectives}
\label{goal}
This experiment is designed to examine various questions concerning the cognitive and reasoning capabilities of LLMs such as Llama2. Specifically, we aim to determine whether such models exhibit characteristics analogous to the dual-system cognitive framework observed in humans (Q1), if self-practice in the absence of Chain of Thought (CoT) guidance enhances reasoning abilities (Q2), whether learning curves indicate improved accuracy with additional examples post self-practice (Q3), if larger models benefit more from self-practice without CoT guidance in terms of performance (Q4), and whether the enhanced reasoning abilities generalized across different reasoning tasks (Q5).

\subsection{Experimental Setting}
To investigate Q1 and Q2 outlined in Section \ref{goal}, we employed untrained LLMs as a baseline to evaluate their efficacy both with and without implementing the CoT method. The few-shot methodology was consistently applied in prompt construction, irrespective of CoT method utilization. Given that the GSM8K dataset's solutions entail reasoning sequences, in instances where the CoT method was not applied, we modified the prompting using GPT-4 \cite{achiam2023gpt} to exclude the reasoning pathway, employing an 8-shot technique. In contrast, for the Reclor and LogiQA2.0 datasets, which naturally lack reasoning pathways in their answers, we engaged GPT-4 to fabricate corresponding reasoning sequences to assess LLMs' proficiency under the CoT paradigm, adopting a 3-shot approach for these datasets. This baseline was then juxtaposed with our CFLLMs framework. To tackle Q3, we experimented with diverse data volumes within the CFLLMs framework to cultivate the LLMs and scrutinized their performance.

In pursuit of Q4, our framework was applied to LLMs of varying sizes, including Vicuna models (7B, 13B, 30B) \cite{zheng2024judging,vicuna2023} and Llama2 models (7B, 13B) \cite{touvron2023llama}. To facilitate deployment on a consumer-grade Nvidia RTX 4090 GPU \cite{NVIDIA2023RTX4090} while minimizing memory usage and inference time, we employed the GPTQ \cite{frantar2022gptq} approach to quantize the models to 4-bit precision. It is important to note that, despite the ability of 7B-sized models to operate on consumer-grade GPUs without quantization, we opted for 4-bit quantization across all models to maintain a uniform comparison scale and minimize quantization errors.

To address the variability in dataset sizes and their potential influence on experimental results, we standardized our approach by extracting a consistent sample of 1000 data entries from each dataset to form the training set for the LLMs' self-practice. An additional 1000 data entries were selected to comprise the test set. To maintain experimental uniformity, each data entry within these subsets was numbered. For each experiment, we consistently used the first $n$ numbered data entries, with $n$ representing the requisite volume of data for the specific experimental conditions.

For Q5, we selected datasets encompassing various reasoning tasks, such as GSM8K \cite{cobbe2021training}, ReClor \cite{yu2020reclor}, and LogiQA2.0 \cite{liu2020logiqa,liu2023logiqa}. GSM8K comprises over 8,000 quality elementary mathematics problems crafted by human authors to assess arithmetic reasoning in LLMs. ReClor features questions from logical reasoning sections of standardized tests like the GMAT and LSAT, challenging the LLMs' critical thinking and complex logical reasoning skills. LogiQA2.0, based on questions from the Chinese civil service exam translated and validated by professional translators and human experts, evaluates the LLMs' capacity for generalizing natural language reasoning.

\subsection{Results}
We conducted experiments across a variety of LLMs, differing in type and size, as well as on diverse datasets, to evaluate their reasoning capabilities. This evaluation was based on the mean answer accuracy derived from five experimental trials, detailed in Table \ref{table1}. It is important to note that the figures succeeding ``CFLLMs'' in the table signify the volume of data utilized for the LLMs' self-practice. The underscored values denote the peak accuracy attained with this methodology for the consistent model and dataset, whereas the bolded values represent the maximum accuracy achieved without employing the CoT method to prompt incremental reasoning, compelling the model to directly generate answers. Red-highlighted numbers in the table reveal that our framework, under the given experimental conditions, did not improve but rather diminished performance.\par

With this groundwork, we can address Q1 and Q2 introduced in Section \ref{goal}. The implementation of CoT markedly influences the models' reasoning proficiency on tasks that entail natural language inference, such as reading comprehension and logical deduction. For instance, on the LogiQA2.0 dataset, the accuracy rates for smaller models like Llama2-7B and Vicuna-7B plummet to near zero in the absence of CoT. However, the deployment of the CogniDual Framework, has resulted in a substantial enhancement of performance without CoT. Despite the models' reasoning accuracy not equalling that of CoT use, their ability to intuitively respond to certain questions suggests an inherent decision-making logic akin to the human dual-system cognitive framework. This insight indicates the potential for transforming System 2 capabilities into System 1 through sustained practice, thereby bolstering the LLMs' rapid response to specific queries and diminishing the time and computational resources required for reasoning. \par

Moreover, we observed a negligible improvement from the CogniDual Framework on the GSM8K dataset, attributed to the models' propensity for step-by-step reasoning even when instructed to directly answer. The prevalence of LLMs producing answers with comprehensive derivations is likely due to task contamination, as postulated by Liu et al. \cite{li2023task}, where mathematical problems are consistently presented with accompanying detailed solutions throughout the training phase. Our framework aims to enhance the System 1 capabilities of LLMs, rather than augment System 2 directly. Consequently, we can deduce from Q5 that only tasks exhibiting a substantial discrepancy in accuracy between CoT usage and non-usage enable LLMs to advance their internalized reasoning abilities through self-practice. \par

For Q3 and Q4, the results in Table \ref{table1} indicate that, in general, an increase in additional examples correlates with a more pronounced enhancement in the LLMs' reasoning abilities without CoT, achieved through self-practice. Larger models require fewer examples to approach their System 1 capacity ceiling; beyond this point, further example data yield minimal benefits. This finding suggests that larger models are more adept at leveraging limited data to improve performance without CoT guidance through self-practice, aligning with the research by Jaimovitch et al. \cite{jaimovitch2021think}.\par

\section{Conclusion}
This study explores the dual cognitive characteristics of LLMs. Our experimental results indicate that once LLMs internalize CoT reasoning through self-training, they can retain CoT-enhanced problem-solving abilities even without CoT prompts. This finding supports the hypothesis that, with appropriate training, LLMs can convert complex, deliberative System 2 reasoning into faster, more intuitive System 1-like responses. Leveraging this property, we designed a self-training framework to reduce the cognitive load of LLM reasoning. Despite these advancements, further research is necessary to address the study's limitations, including examining how this framework influences the cognitive processing preferences of LLMs.

\clearpage

\bibliographystyle{IEEEtran}
\bibliography{citation}

\begin{thebibliography}{10}
\providecommand{\url}[1]{#1}
\csname url@samestyle\endcsname
\providecommand{\newblock}{\relax}
\providecommand{\bibinfo}[2]{#2}
\providecommand{\BIBentrySTDinterwordspacing}{\spaceskip=0pt\relax}
\providecommand{\BIBentryALTinterwordstretchfactor}{4}
\providecommand{\BIBentryALTinterwordspacing}{\spaceskip=\fontdimen2\font plus
\BIBentryALTinterwordstretchfactor\fontdimen3\font minus \fontdimen4\font\relax}
\providecommand{\BIBforeignlanguage}[2]{{%
\expandafter\ifx\csname l@#1\endcsname\relax
\typeout{** WARNING: IEEEtran.bst: No hyphenation pattern has been}%
\typeout{** loaded for the language `#1'. Using the pattern for}%
\typeout{** the default language instead.}%
\else
\language=\csname l@#1\endcsname
\fi
#2}}
\providecommand{\BIBdecl}{\relax}
\BIBdecl

\bibitem{evans2013dual}
J.~S.~B. Evans and K.~E. Stanovich, ``Dual-process theories of higher cognition: Advancing the debate,'' \emph{Perspectives on psychological science}, vol.~8, no.~3, pp. 223--241, 2013.

\bibitem{stanovich2008relative}
K.~E. Stanovich and R.~F. West, ``On the relative independence of thinking biases and cognitive ability.'' \emph{Journal of personality and social psychology}, vol.~94, no.~4, p. 672, 2008.

\bibitem{kahneman2011thinking}
D.~Kahneman, \emph{Thinking, fast and slow}.\hskip 1em plus 0.5em minus 0.4em\relax macmillan, 2011.

\bibitem{couldry2020costs}
N.~Couldry and U.~A. Mejias, \emph{The costs of connection: How data is colonizing human life and appropriating it for capitalism}.\hskip 1em plus 0.5em minus 0.4em\relax Stanford University Press, 2020.

\bibitem{guest2021computational}
O.~Guest and A.~E. Martin, ``How computational modeling can force theory building in psychological science,'' \emph{Perspectives on Psychological Science}, vol.~16, no.~4, pp. 789--802, 2021.

\bibitem{pertwee2022epidemic}
E.~Pertwee, C.~Simas, and H.~J. Larson, ``An epidemic of uncertainty: rumors, conspiracy theories and vaccine hesitancy,'' \emph{Nature medicine}, vol.~28, no.~3, pp. 456--459, 2022.

\bibitem{balland2022reprint}
P.-A. Balland, T.~Broekel, D.~Diodato, E.~Giuliani, R.~Hausmann, N.~O'Clery, and D.~Rigby, ``Reprint of the new paradigm of economic complexity,'' \emph{Research Policy}, vol.~51, no.~8, p. 104568, 2022.

\bibitem{post2020advancing}
C.~Post, R.~Sarala, C.~Gatrell, and J.~E. Prescott, ``Advancing theory with review articles,'' \emph{Journal of Management Studies}, vol.~57, no.~2, pp. 351--376, 2020.

\bibitem{vaswani2017attention}
\BIBentryALTinterwordspacing
A.~Vaswani, N.~Shazeer, N.~Parmar, J.~Uszkoreit, L.~Jones, A.~N. Gomez, L.~Kaiser, and I.~Polosukhin, ``Attention is all you need,'' in \emph{Advances in Neural Information Processing Systems 30: Annual Conference on Neural Information Processing Systems 2017, December 4-9, 2017, Long Beach, CA, {USA}}, I.~Guyon, U.~von Luxburg, S.~Bengio, H.~M. Wallach, R.~Fergus, S.~V.~N. Vishwanathan, and R.~Garnett, Eds., 2017, pp. 5998--6008. [Online]. Available: \url{https://proceedings.neurips.cc/paper/2017/hash/3f5ee243547dee91fbd053c1c4a845aa-Abstract.html}
\BIBentrySTDinterwordspacing

\bibitem{achiam2023gpt}
\BIBentryALTinterwordspacing
J.~Achiam, S.~Adler, S.~Agarwal, L.~Ahmad, I.~Akkaya, F.~L. Aleman, D.~Almeida, J.~Altenschmidt, S.~Altman, S.~Anadkat \emph{et~al.}, ``Gpt-4 technical report,'' \emph{ArXiv preprint}, vol. abs/2303.08774, 2023. [Online]. Available: \url{https://arxiv.org/abs/2303.08774}
\BIBentrySTDinterwordspacing

\bibitem{ouyang2022training}
L.~Ouyang, J.~Wu, X.~Jiang, D.~Almeida, C.~Wainwright, P.~Mishkin, C.~Zhang, S.~Agarwal, K.~Slama, A.~Ray \emph{et~al.}, ``Training language models to follow instructions with human feedback,'' \emph{Advances in Neural Information Processing Systems}, vol.~35, pp. 27\,730--27\,744, 2022.

\bibitem{wang2022self}
\BIBentryALTinterwordspacing
Y.~Wang, Y.~Kordi, S.~Mishra, A.~Liu, N.~A. Smith, D.~Khashabi, and H.~Hajishirzi, ``Self-instruct: Aligning language model with self generated instructions,'' \emph{ArXiv preprint}, vol. abs/2212.10560, 2022. [Online]. Available: \url{https://arxiv.org/abs/2212.10560}
\BIBentrySTDinterwordspacing

\bibitem{taori2023alpaca}
R.~Taori, I.~Gulrajani, T.~Zhang, Y.~Dubois, X.~Li, C.~Guestrin, P.~Liang, and T.~B. Hashimoto, ``Alpaca: A strong, replicable instruction-following model,'' \emph{Stanford Center for Research on Foundation Models. https://crfm. stanford. edu/2023/03/13/alpaca. html}, vol.~3, no.~6, p.~7, 2023.

\bibitem{deng2024promoting}
\BIBentryALTinterwordspacing
Y.~Deng, X.~Qiu, X.~Tan, J.~Pan, C.~Jue, Z.~Fang, Y.~Xu, W.~Chu, and Y.~Qi, ``Promoting equality in large language models: Identifying and mitigating the implicit bias based on bayesian theory,'' \emph{ArXiv preprint}, vol. abs/2408.10608, 2024. [Online]. Available: \url{https://arxiv.org/abs/2408.10608}
\BIBentrySTDinterwordspacing

\bibitem{wei2022chain}
J.~Wei, X.~Wang, D.~Schuurmans, M.~Bosma, F.~Xia, E.~Chi, Q.~V. Le, D.~Zhou \emph{et~al.}, ``Chain-of-thought prompting elicits reasoning in large language models,'' \emph{Advances in Neural Information Processing Systems}, vol.~35, pp. 24\,824--24\,837, 2022.

\bibitem{kojima2022large}
T.~Kojima, S.~S. Gu, M.~Reid, Y.~Matsuo, and Y.~Iwasawa, ``Large language models are zero-shot reasoners,'' \emph{Advances in neural information processing systems}, vol.~35, pp. 22\,199--22\,213, 2022.

\bibitem{hu2021lora}
\BIBentryALTinterwordspacing
E.~J. Hu, Y.~Shen, P.~Wallis, Z.~Allen{-}Zhu, Y.~Li, S.~Wang, L.~Wang, and W.~Chen, ``Lora: Low-rank adaptation of large language models,'' in \emph{The Tenth International Conference on Learning Representations, {ICLR} 2022, Virtual Event, April 25-29, 2022}.\hskip 1em plus 0.5em minus 0.4em\relax OpenReview.net, 2022. [Online]. Available: \url{https://openreview.net/forum?id=nZeVKeeFYf9}
\BIBentrySTDinterwordspacing

\bibitem{clavie2023large}
B.~Clavi{\'e}, A.~Ciceu, F.~Naylor, G.~Souli{\'e}, and T.~Brightwell, ``Large language models in the workplace: A case study on prompt engineering for job type classification,'' in \emph{International Conference on Applications of Natural Language to Information Systems}.\hskip 1em plus 0.5em minus 0.4em\relax Springer, 2023, pp. 3--17.

\bibitem{huang2022context}
\BIBentryALTinterwordspacing
Y.~Huang, Y.~Chen, Z.~Yu, and K.~McKeown, ``In-context learning distillation: Transferring few-shot learning ability of pre-trained language models,'' \emph{ArXiv preprint}, vol. abs/2212.10670, 2022. [Online]. Available: \url{https://arxiv.org/abs/2212.10670}
\BIBentrySTDinterwordspacing

\bibitem{agarwal2023gkd}
\BIBentryALTinterwordspacing
R.~Agarwal, N.~Vieillard, P.~Stanczyk, S.~Ramos, M.~Geist, and O.~Bachem, ``Gkd: Generalized knowledge distillation for auto-regressive sequence models,'' \emph{ArXiv preprint}, vol. abs/2306.13649, 2023. [Online]. Available: \url{https://arxiv.org/abs/2306.13649}
\BIBentrySTDinterwordspacing

\bibitem{li2022explanations}
\BIBentryALTinterwordspacing
S.~Li, J.~Chen, Y.~Shen, Z.~Chen, X.~Zhang, Z.~Li, H.~Wang, J.~Qian, B.~Peng, Y.~Mao \emph{et~al.}, ``Explanations from large language models make small reasoners better,'' \emph{ArXiv preprint}, vol. abs/2210.06726, 2022. [Online]. Available: \url{https://arxiv.org/abs/2210.06726}
\BIBentrySTDinterwordspacing

\bibitem{ho2022large}
\BIBentryALTinterwordspacing
N.~Ho, L.~Schmid, and S.-Y. Yun, ``Large language models are reasoning teachers,'' \emph{ArXiv preprint}, vol. abs/2212.10071, 2022. [Online]. Available: \url{https://arxiv.org/abs/2212.10071}
\BIBentrySTDinterwordspacing

\bibitem{fu2023specializing}
\BIBentryALTinterwordspacing
Y.~Fu, H.~Peng, L.~Ou, A.~Sabharwal, and T.~Khot, ``Specializing smaller language models towards multi-step reasoning,'' \emph{ArXiv preprint}, vol. abs/2301.12726, 2023. [Online]. Available: \url{https://arxiv.org/abs/2301.12726}
\BIBentrySTDinterwordspacing

\bibitem{shridhar2023distilling}
K.~Shridhar, A.~Stolfo, and M.~Sachan, ``Distilling reasoning capabilities into smaller language models,'' in \emph{Findings of the Association for Computational Linguistics: ACL 2023}, 2023, pp. 7059--7073.

\bibitem{hsieh2023distilling}
\BIBentryALTinterwordspacing
C.-Y. Hsieh, C.-L. Li, C.-K. Yeh, H.~Nakhost, Y.~Fujii, A.~Ratner, R.~Krishna, C.-Y. Lee, and T.~Pfister, ``Distilling step-by-step! outperforming larger language models with less training data and smaller model sizes,'' \emph{ArXiv preprint}, vol. abs/2305.02301, 2023. [Online]. Available: \url{https://arxiv.org/abs/2305.02301}
\BIBentrySTDinterwordspacing

\bibitem{zheng2024judging}
L.~Zheng, W.-L. Chiang, Y.~Sheng, S.~Zhuang, Z.~Wu, Y.~Zhuang, Z.~Lin, Z.~Li, D.~Li, E.~Xing \emph{et~al.}, ``Judging llm-as-a-judge with mt-bench and chatbot arena,'' \emph{Advances in Neural Information Processing Systems}, vol.~36, 2024.

\bibitem{vicuna2023}
W.-L. Chiang, Z.~Li, Z.~Lin, Y.~Sheng, Z.~Wu, H.~Zhang, L.~Zheng, S.~Zhuang, Y.~Zhuang, J.~E. Gonzalez, I.~Stoica, and E.~P. Xing, ``Vicuna: An open-source chatbot impressing gpt-4 with 90\%* chatgpt quality,'' 2023.

\bibitem{touvron2023llama}
\BIBentryALTinterwordspacing
H.~Touvron, L.~Martin, K.~Stone, P.~Albert, A.~Almahairi, Y.~Babaei, N.~Bashlykov, S.~Batra, P.~Bhargava, S.~Bhosale \emph{et~al.}, ``Llama 2: Open foundation and fine-tuned chat models,'' \emph{ArXiv preprint}, vol. abs/2307.09288, 2023. [Online]. Available: \url{https://arxiv.org/abs/2307.09288}
\BIBentrySTDinterwordspacing

\bibitem{NVIDIA2023RTX4090}
NVIDIA, ``Nvidia geforce rtx 4090,'' 2023, accessed: 2024-04-30.

\bibitem{frantar2022gptq}
\BIBentryALTinterwordspacing
E.~Frantar, S.~Ashkboos, T.~Hoefler, and D.~Alistarh, ``Gptq: Accurate post-training quantization for generative pre-trained transformers,'' \emph{ArXiv preprint}, vol. abs/2210.17323, 2022. [Online]. Available: \url{https://arxiv.org/abs/2210.17323}
\BIBentrySTDinterwordspacing

\bibitem{cobbe2021training}
\BIBentryALTinterwordspacing
K.~Cobbe, V.~Kosaraju, M.~Bavarian, M.~Chen, H.~Jun, L.~Kaiser, M.~Plappert, J.~Tworek, J.~Hilton, R.~Nakano \emph{et~al.}, ``Training verifiers to solve math word problems,'' \emph{ArXiv preprint}, vol. abs/2110.14168, 2021. [Online]. Available: \url{https://arxiv.org/abs/2110.14168}
\BIBentrySTDinterwordspacing

\bibitem{yu2020reclor}
\BIBentryALTinterwordspacing
W.~Yu, Z.~Jiang, Y.~Dong, and J.~Feng, ``Reclor: {A} reading comprehension dataset requiring logical reasoning,'' in \emph{8th International Conference on Learning Representations, {ICLR} 2020, Addis Ababa, Ethiopia, April 26-30, 2020}.\hskip 1em plus 0.5em minus 0.4em\relax OpenReview.net, 2020. [Online]. Available: \url{https://openreview.net/forum?id=HJgJtT4tvB}
\BIBentrySTDinterwordspacing

\bibitem{liu2020logiqa}
\BIBentryALTinterwordspacing
J.~Liu, L.~Cui, H.~Liu, D.~Huang, Y.~Wang, and Y.~Zhang, ``Logiqa: {A} challenge dataset for machine reading comprehension with logical reasoning,'' in \emph{Proceedings of the Twenty-Ninth International Joint Conference on Artificial Intelligence, {IJCAI} 2020}, C.~Bessiere, Ed.\hskip 1em plus 0.5em minus 0.4em\relax ijcai.org, 2020, pp. 3622--3628. [Online]. Available: \url{https://doi.org/10.24963/ijcai.2020/501}
\BIBentrySTDinterwordspacing

\bibitem{liu2023logiqa}
H.~Liu, J.~Liu, L.~Cui, Z.~Teng, N.~Duan, M.~Zhou, and Y.~Zhang, ``Logiqa 2.0—an improved dataset for logical reasoning in natural language understanding,'' \emph{IEEE/ACM Transactions on Audio, Speech, and Language Processing}, 2023.

\bibitem{li2023task}
\BIBentryALTinterwordspacing
C.~Li and J.~Flanigan, ``Task contamination: Language models may not be few-shot anymore,'' \emph{ArXiv preprint}, vol. abs/2312.16337, 2023. [Online]. Available: \url{https://arxiv.org/abs/2312.16337}
\BIBentrySTDinterwordspacing

\bibitem{jaimovitch2021think}
\BIBentryALTinterwordspacing
G.~Jaimovitch{-}Lopez, D.~C. Falc{\'{o}}n, C.~Ferri, and J.~Hern{\'{a}}ndez{-}Orallo, ``Think big, teach small: Do language models distil occam's razor?'' in \emph{Advances in Neural Information Processing Systems 34: Annual Conference on Neural Information Processing Systems 2021, NeurIPS 2021, December 6-14, 2021, virtual}, M.~Ranzato, A.~Beygelzimer, Y.~N. Dauphin, P.~Liang, and J.~W. Vaughan, Eds., 2021, pp. 1610--1623. [Online]. Available: \url{https://proceedings.neurips.cc/paper/2021/hash/0cd6a652ed1f7811192db1f700c8f0e7-Abstract.html}
\BIBentrySTDinterwordspacing

\end{thebibliography}

\end{document}